# Not quite unreasonable effectiveness of machine learning algorithms

Egor Illarionov[1,2,*], Roman Khudorozhkov[1]

*Abstract*—State-of-the-art machine learning algorithms demonstrate close to absolute performance in selected challenges. We provide arguments that the reason can be in low variability of the samples and high effectiveness in learning typical patterns. Due to this fact, standard performance metrics do not reveal model capacity and new metrics are required for the better understanding of state-of-the-art.

## I. INTRODUCTION

There are more and more papers that report state-of-the-art results in various machine learning challenges. For example, accuracy of handwritten digits classification based on the MNIST dataset (LeCun and Cortes 2010) reaches 99.69% in Liang and Hu (2015), 99.77% in Cireşan, Meier, and Schmidhuber (2012) and 99.79% in Wan et al. (2013). However, there are only few articles which investigate datasets and consider factors which simplify or complicate model learning. For example, Li (2006) considers principal subsets which are the best approximations of the dataset. Practical implementation of this idea is based on SVM classification algorithm and this fact reduces a variety of problems where the approach can be applied. On the other hand, Zubek and Plewczynski (2016) investigate data complexity based on the underlying probability distribution and Hellinger distance. This approach seems reasonable when items in the dataset are represented by feature vectors, while for datasets with images the implementation is not straightforward.

An alternative approach considers the complexity of the classifier decision boundary and characterizes it by a number of geometrical descriptors (T. K. Ho and M. Basu 2002; T. Ho, Mitra Basu, and Hiu Chung Law 2006). However, geometrical considerations become complicated in high-dimension spaces. Rolnick et al. (2017) investigate the influence of mislabeling on classification performance and demonstrate that results remain robust. In the paper Szegedy, Zaremba, et al. (2013) one analyzes the robustness to adversarial examples.

We suggest another point of view. Reaching high performance can be possible due to the high similarity between train and test parts of the dataset. This idea is supported by the fact that only 10% of the MNIST train samples ensure over 97% of accuracy for typical neural network architectures. The high similarity could be understood as limited number of handwriting styles or patterns existing in the dataset. Having learned them, the model can predict any digit from the dataset.

Of course, we do not know the real number of handwriting styles in the dataset as it is remain unknown which person wrote which digit. However, there are datasets where we can restore this analogy. As such we consider a medical dataset of electrocardiogram (ECG) signals, namely MIT-BIH Atrial Fibrillation dataset (Goldberger et al. 2000). It consists of 10-hours ECG samples from 23 patients. Heart rhythm are not homogeneous along signals and contain segments labeled as normal rhythm and atrial fibrillation.

By cropping a large number of short segments from each signal we obtain a large dataset of labeled signals, which are randomly split into train and test parts. Although all segments in train and test parts are unique, there is a very limited number of patterns, each of which corresponds to a unique patient (although one patient might have several patterns). We will show that the model, which can classify segments with close to 1 accuracy, in fact also accumulates information about each patient.

## II. COMPLEXITY OF DATASETS

The similarity between samples, which we want to investigate, is closely related to the notion of sample complexity defined in the statistical learning theory (Hastie, Tibshirani, and Friedman 2001) and to the Kolmogorov complexity (Kolmogorov 1998).

Indeed, sample complexity is defined (informally) as the minimal number of samples which is required for learning the model with any given error rate. If we could find a model with low sample complexity for a given dataset, we could assume an existence of high internal similarity between samples.

Similarity can also be interpreted as an amount of information contained in a given sample (or set) relative to another sample (or set). This can be written as $KS(x|y)$, where $KS$ is for the Kolmogorov complexity.

Although both points of view are theoretically well-defined, their practical implementation is marginal, to say the least. Moreover, $KS$ is uncomputable. To overcome this fact, we will estimate the similarity between two sets of samples as similarity in models evaluation considering one set as a train set, and another one as a test set. If we can find a model which requires a low number of train samples to reach high performance on the test, this would indicate both low sample complexity and low relative complexity of both sets.

## III. COMPLEXITY OF THE MNIST DATASET

The MNIST dataset is often exploited for demonstration of results in machine learning. For example, it was mentioned in important deep learning papers He et al. (2015)

[1]Moscow IT Department, Moscow, Russia
[2]Moscow State University, Moscow, Russia
[*]Corresponding author: e.illarionov@analysiscenter.ru

and Goodfellow et al. (2014). However, it is still an open question to what extent results based on this dataset can be representative.

Let's recall some facts on the MNIST dataset. It is composed of 70K grayscale images of digits, each image has a size of $28 \times 28$ pixels. The dataset is split into train and test in proportion 6 to 1.

We will train two classic neural network architectures, ResNet18 (He et al. 2015) and Inception_v1 (Szegedy, Liu, et al. 2014) with default parameters. These models demonstrate high accuracy in various image classification problems, see e.g. Ghanem et al. (2017). For comparison we also consider a simple fully connected (FC) model with 28*28 inputs, 64 neurons in the hidden layer (with relu activation) and 10 outputs (with softmax activation). All the models are implemented and run with the Dataset framework (Kh and al 2017).

Figure 1 shows how the accuracy on the test dataset grows with an increase of the utilized train dataset share.

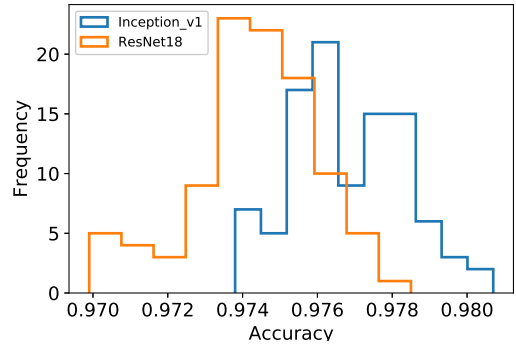

Fig. 2. Model accuracy over 100 random subsets. Each subset is 7% of the train dataset. Blue is for the Inception_v1 model, orange is for the ResNet18 model.

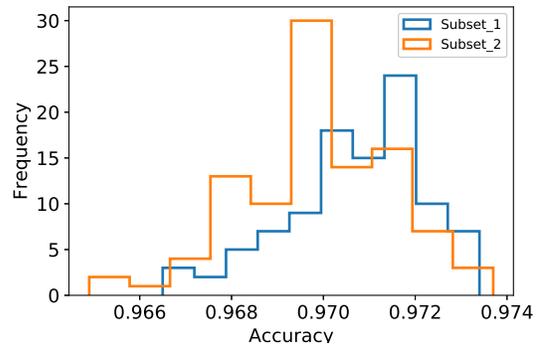

Fig. 3. ResNet18 model accuracy over 100 initialization for 2 different subsets. Subsets are 5% of the train dataset. Orange is for random subset, blue is for subset selected with some criteria.

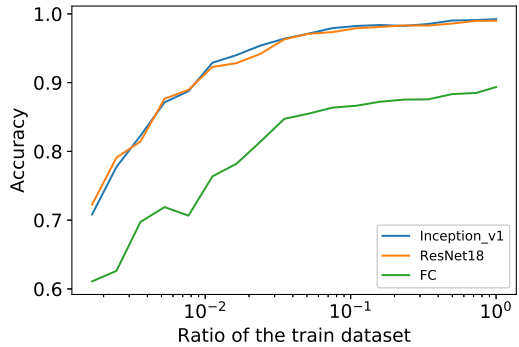

Fig. 1. Model accuracy against percentage of the train dataset utilized. Blue is for the Inception_v1 model, orange is for the ResNet18 model, green is for the FC model. Note that x-axis has logarithmic scale.

We observe that the curve saturates at about 10% of the train dataset. More precisely, the ResNet18 model requires 7% of the train dataset (4.2K samples) to reach 97% accuracy on the test dataset. We validated this on 100 random subsets and accuracy was below 97% only 1 time out of 100 (see figure 2). We obtain even better results for the Inception_v1 model. It shows above 97% accuracy for each of the 100 random subsets.

We noted that for the ResNet18 model we can reduce the percentage of the train size utilized up to 5% and keep the mean accuracy over 97% for different model initializations. However, samples should be selected more carefully. Figure 3 shows that the model accuracy distribution differs if samples are selected at random and if they are chosen with some criteria. In the first case, 41 of 100 realizations show an accuracy of above 97%, in the second case, we obtained already 72 realizations with an accuracy of above 97%. An interesting question is how to select even less samples without the loss in model accuracy.

Note that the FC model, which in general underperforms ResNet18 and Inception_v1, also approaches its maximal accuracy after visiting about 10% of the train dataset (figure 1).

Thus we observe that a small fraction of the dataset can efficiently approximate general distribution, which follow all images in the MNIST. In our opinion, this fact demonstrates a low variability in the dataset. As a consequence we assume that enlarging of the train dataset with new samples (which follow the general distribution) will not provide significant accuracy increase for the FC model. We can say that the low accuracy of the FC model is mostly due to its architecture rather than due to the insufficient number of training samples.

Low variability of the MNIST dataset bring us to the idea that there is a limited number of patterns (handwritings styles) within the digits. The point is that these notions are difficult to formalize and we are unable to validate this idea as there is no information about a person who write each digit. However, we can draw an analogy with the other dataset, where this idea can be easily validated. For this purpose we consider a large dataset of heart signals obtained from small number of patients. Here signal patterns have natural association with patients. For example, Fratini et al. (2015) use this fact for individual identification via electrocardiogram analysis. In our paper, learning to classify signals into, say, normal and abnormal rhythms, we will check, whether the model accumulates information about patients. We discuss this in the next section.

## IV. COMPLEXITY OF THE ATRIAL FIBRILLATION DATASET

We consider the MIT-BIH Atrial Fibrillation database, which is commonly used for the evaluation heart disease detection models. For example, Dash et al. (2009) reported about sensitivity and specificity 94.4% and 95.1% for atrial fibrillation detection, while Johura et al. (2017) achieved sensitivity of 98%, specificity of 98% and accuracy of 95%.

The database consists of about 10-hours ECG signals from 23 patients. Each ECG is represented by two signals which correspond to two physical leads. Only the first lead will be exploited. The ECG signal is divided into segments that represent normal and abnormal heart rhythm. More precisely, we will refer segments, annotated as "AFIB" to one class (class A), while all other labels, which represent mostly normal rhythm, will compose another class (class N). Figures 4 and 5 show examples of both classes.

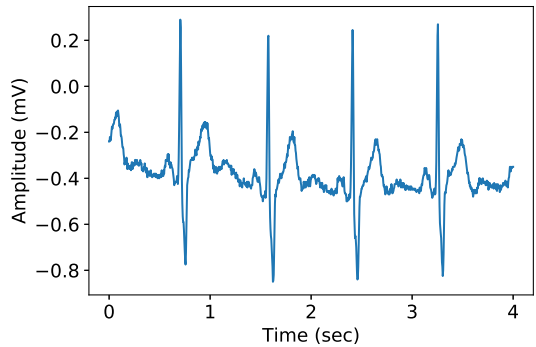

Fig. 4. Example of normal rhythm.

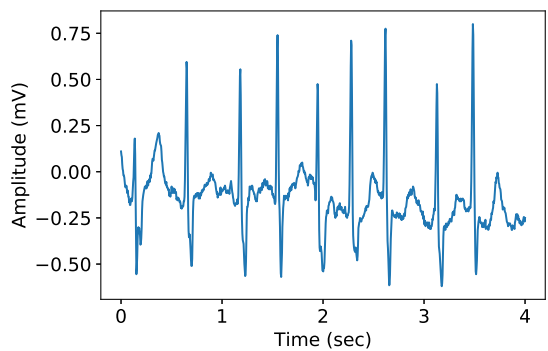

Fig. 5. Example of atrial fibrillation.

Then we split each ECG into short segments of 16 sec long. Note that each segment corresponds to a vector of shape 4000 for a signal sampled at 250 Hz. Segments are labeled by one of two classes following the annotation. Table I shows the number of segments obtained from some patients. Note that the distribution varies substantially. Totally we obtain about 52K segments from which 21K have label A. All the segments are randomly split into train and test datasets in proportion 70 to 30.

TABLE I
DISTRIBUTION OF LABELS FOR SOME PATIENTS

| Patient_id | Number of A segments | Number of N segments |
|---|---|---|
| 04015 | 13 | 2282 |
| 04746 | 1219 | 1077 |
| 04936 | 1642 | 623 |
| 07162 | 2301 | 0 |
| 08219 | 476 | 1784 |

Note that heartbeats from the same patient can appear in the train and test datasets. This heartbeat division scheme is called intra-patient (Lannoy et al. 2012). An alternative approach (inter-patient) uses different patients in train and test datasets. A detailed comparison of both schemes can be found in S. Luz et al. (2016). While the second scheme (inter-patient) could be intuitively assumed as more realistic and fair, in our opinion, it only shifts the problem of similarity of heartbeats within one patient to the similarity of patients within the database and does not provide convincing arguments that the problem is thus diminished. An approach developed in this paper can be applied for systematic investigation of the inter-patient scheme.

For classification we will train a neural network model implemented in the CardIO framework (Khudorozhkov et al. 2017). Generally, there are many papers which analyze ECGs with neural networks (see, e.g. recent paper Rajpurkar et al. (2017)). An interesting feature of the considered model is that it investigates the frequency domain of ECG signals. The model (called FFT model) consists of several blocks (see figure 6). In the beginning it applies a set of 1D convolutions to the input signal. After 4 convolutions and max-pooling layers the fast Fourier transform is applied. All Fourier spectrum are then stacked into one 2D image, which is fed into a sequence of inception blocks and max-pooling layers. In the end we apply global max-pooling and two dense layers with 8 and 2 neurons and obtain output vector of shape 2.

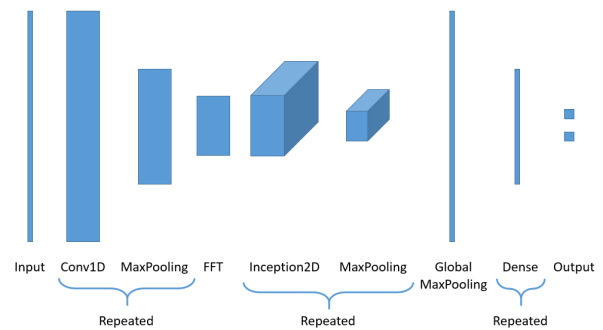

Fig. 6. Architecture of the FFT model from the CardIO framework.

As well as in the previous section, we will increase a proportion of the train dataset utilized, train the model and consider the performance on the test dataset. The model is trained over 150 epochs with macro F1 score as metric, which is averaged F1 score for each class. Figure 7 shows that F1 score saturates at only 25% of the train dataset. This fact supports the idea of high similarity between segments.

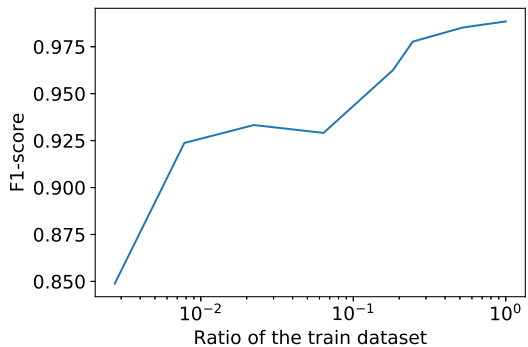

Fig. 7. F1 score against percentage of the train dataset utilized. Note that x-axis has a logarithmic scale.

Note that the highest score is 0.988. It outperforms many previously reported results on atrial fibrillation detection, e.g. recent Johura et al. (2017).

To demonstrate that the model accumulates information about patients we consider the global max-pooling layer. This layer has output size 40. For each segment we calculate output from this layer and thus obtain feature vectors. The target is an id of the patient, from which ECG the segment was sampled. Feature vectors and targets obtained from the train dataset are used to train the Random forest classification algorithms (T. K. Ho 1995). Table II shows the classification report on the test dataset. We observe the overall accuracy is 95%.

TABLE II
CLASSIFICATION REPORT

| Patient_id | Precision | Recall | F1-score |
|---|---|---|---|
| 04015 | 0.93 | 0.96 | 0.94 |
| 04043 | 0.98 | 0.97 | 0.98 |
| 04048 | 0.97 | 0.97 | 0.97 |
| 04126 | 0.92 | 0.96 | 0.94 |
| 04746 | 0.92 | 0.90 | 0.91 |
| 04908 | 0.97 | 0.99 | 0.98 |
| 04936 | 0.97 | 0.97 | 0.97 |
| 05091 | 0.94 | 0.97 | 0.95 |
| 05121 | 0.89 | 0.93 | 0.91 |
| 05261 | 0.95 | 0.90 | 0.92 |
| 06426 | 0.93 | 0.85 | 0.89 |
| 06453 | 0.98 | 0.96 | 0.97 |
| 06995 | 0.97 | 0.97 | 0.97 |
| 07162 | 1.00 | 0.99 | 1.00 |
| 07859 | 0.99 | 0.99 | 0.99 |
| 07879 | 0.89 | 0.91 | 0.90 |
| 07910 | 0.96 | 0.99 | 0.97 |
| 08215 | 0.98 | 0.97 | 0.97 |
| 08219 | 0.94 | 0.93 | 0.94 |
| 08378 | 0.94 | 0.94 | 0.94 |
| 08405 | 0.89 | 0.88 | 0.89 |
| 08434 | 0.97 | 0.96 | 0.96 |
| 08455 | 0.89 | 0.89 | 0.89 |
| avg / total | 0.95 | 0.95 | 0.95 |

The fact that we can successfully identify patients based on their heart signal feature vectors has several consequences. First, within each signal there exists a limited number of patterns. Second, patterns do not vary a lot within signals. Third, patterns are patient unique.

These observations support the idea that high accuracy in classification of heart diseases from MIT-BIH Atrial Fibrillation database is possible due to low variability of signal patterns within each signal.

## V. CONCLUSION

This article investigates the question why it is possible to reach high performance results with machine learning algorithms on some datasets. We suggest an idea that it is possible due to low variability of patterns that exist in datasets and ability on modern neural network architectures to efficiently learn these patterns.

We considered the MNIST dataset and showed that only 10% of the train dataset can ensure over 97% of accuracy on the test dataset. It supports the idea of the low variability in the dataset. Generalizing, one could assume that there exists a limited number of handwriting styles, which do not allow digits to vary a lot. It is difficult to verify this hypothesis on the MNIST dataset, however, we can draw an analogy with the dataset of heart signals, where patterns can be associated with the patient, from which the ECG signal is sampled.

We investigated the MIT-BIH Atrial Fibrillation dataset and demonstrated that the FFT model from CardIO framework reaches almost 99% of performance in classification of atrial fibrillation from heart signals. At the same time, 25% of the train dataset ensures over 97% in model evaluation. This supports the idea of low variability in the data. Moreover, signal patterns can be associated with patients. It was demonstrated by the fact that hidden representation of signals can be considered as feature vectors for accurate patient identification.

This brings us to the conclusion that standard metrics do not allow to evaluate modern machine learning algorithms efficiently and recognize state-of-the-art solutions. We suggest to introduce new metrics that better reveal model capacity. This investigation is out of the scope of this article, however, we can suggest some directions:

- investigation of model variance with respect to random initialization, percentage of train dataset utilized, representativity of the train dataset;
- selecting samples where the model fails and understanding the reasons;
- recovering additional knowledge about the subject area, which the model provides.

In our opinion, the progress in elaboration of machine learning algorithms should be followed by the elaboration of new evaluation methods, which are able to reveal its hidden potential. Unfortunately, most ideas remain only in theory and do not have practical application. We believe that the need of new quantitative metrics will become more and more clear, since more and more algorithms demonstrate almost similar results according to standard evaluation schemes.

## REFERENCES

Cireşan, D., U. Meier, and J. Schmidhuber (2012). "Multi-column Deep Neural Networks for Image Classification". In: *ArXiv e-prints*. arXiv: 1202.2745 [cs.CV].

Dash, S. et al. (2009). "Automatic Real Time Detection of Atrial Fibrillation". In: *Annals of Biomedical Engineering* 37, pp. 1701–1709.

Fratini, Antonio et al. (2015). "Individual identification via electrocardiogram analysis". In: *Biomedical engineering online*.

Ghanem, B. et al. (2017). "ActivityNet Challenge 2017 Summary". In: *ArXiv e-prints*. arXiv: 1710.08011 [cs.CV].

Goldberger, A. L. et al. (2000). "PhysioBank, PhysioToolkit, and PhysioNet: Components of a New Research Resource for Complex Physiologic Signals". In: *Circulation* 101.23. Circulation Electronic Pages: http://circ.ahajournals.org/content/101/23/e215.full PMID:1085218; doi: 10.1161/01.CIR.101.23.e215, e215–e220.

Goodfellow, I. J. et al. (2014). "Generative Adversarial Networks". In: *ArXiv e-prints*. arXiv: 1406.2661 [stat.ML].

Hastie, Trevor, Robert Tibshirani, and Jerome Friedman (2001). *The Elements of Statistical Learning*. Springer Series in Statistics. New York, NY, USA: Springer New York Inc.

He, K. et al. (2015). "Deep Residual Learning for Image Recognition". In: *ArXiv e-prints*. arXiv: 1512.03385 [cs.CV].

Ho, Tin Kam (1995). "Random decision forests". In: *Proceedings of 3rd International Conference on Document Analysis and Recognition*. Vol. 1, 278–282 vol.1. DOI: 10.1109/ICDAR.1995.598994.

Ho, Tin Kam and M. Basu (2002). "Complexity measures of supervised classification problems". In: *IEEE Transactions on Pattern Analysis and Machine Intelligence* 24.3, pp. 289–300. ISSN: 0162-8828. DOI: 10.1109/34.990132.

Ho, Tin, Mitra Basu, and Martin Hiu Chung Law (2006). "Measures of Geometrical Complexity in Classification Problems". In: *Data Complexity in Pattern Recognition, Advanced Information and Knowledge Processing*. Vol. 16, pp. 1–23.

Johura, F. T. et al. (2017). "ECG signal for artrial fibrillation detection". In: *2017 International Conference on Electrical, Computer and Communication Engineering (ECCE)*, pp. 928–934. DOI: 10.1109/ECACE.2017.7913036.

Kh, Roman and et al (2017). *Dataset library for fast ML workflows*. DOI: 10.5281/zenodo.1041203. URL: https://doi.org/10.5281/zenodo.1041203.

Khudorozhkov, R. et al. (2017). *CardIO library for deep research of heart signals*. DOI: 10.5281/zenodo.1156085. URL: https://doi.org/10.5281/zenodo.1156085.

Kolmogorov, A.N. (1998). "On tables of random numbers". In: *Theoretical Computer Science* 207.2, pp. 387–395. ISSN: 0304-3975. DOI: https://doi.org/10.1016/S0304-3975(98)00075-9. URL: http://www.sciencedirect.com/science/article/pii/S0304397598000759.

Lannoy, G. de et al. (2012). "Weighted Conditional Random Fields for Supervised Interpatient Heartbeat Classification". In: *IEEE Transactions on Biomedical Engineering* 59.1, pp. 241–247. ISSN: 0018-9294. DOI: 10.1109/TBME.2011.2171037.

LeCun, Yann and Corinna Cortes (2010). "MNIST handwritten digit database". In: URL: http://yann.lecun.com/exdb/mnist/.

Li, Ling (2006). "Data Complexity in Machine Learning and Novel Classification Algorithms". AAI3235581. PhD thesis. Pasadena, CA, USA. ISBN: 978-0-542-89256-1.

Liang, Ming and Xiaolin Hu (2015). "Recurrent convolutional neural network for object recognition". In: *2015 IEEE Conference on Computer Vision and Pattern Recognition (CVPR)*, pp. 3367–3375. DOI: 10.1109/CVPR.2015.7298958.

Rajpurkar, P. et al. (2017). "Cardiologist-Level Arrhythmia Detection with Convolutional Neural Networks". In: *ArXiv e-prints*. arXiv: 1707.01836 [cs.CV].

Rolnick, D. et al. (2017). "Deep Learning is Robust to Massive Label Noise". In: *ArXiv e-prints*. arXiv: 1705.10694 [cs.LG].

S. Luz, Eduardo Jose da et al. (2016). "ECG-based heartbeat classification for arrhythmia detection: A survey". In: *Computer Methods and Programs in Biomedicine* 127, pp. 144–164. ISSN: 0169-2607. DOI: https://doi.org/10.1016/j.cmpb.2015.12.008. URL: http://www.sciencedirect.com/science/article/pii/S0169260715003314.

Szegedy, C., W. Liu, et al. (2014). "Going Deeper with Convolutions". In: *ArXiv e-prints*. arXiv: 1409.4842 [cs.CV].

Szegedy, C., W. Zaremba, et al. (2013). "Intriguing properties of neural networks". In: *ArXiv e-prints*. arXiv: 1312.6199 [cs.CV].

Wan, Li et al. (2013). "Regularization of Neural Networks using DropConnect". In: *Proceedings of the 30th International Conference on Machine Learning*. Ed. by Sanjoy Dasgupta and David McAllester. Vol. 28. Proceedings of Machine Learning Research 3. Atlanta, Georgia, USA: PMLR, pp. 1058–1066. URL: http://proceedings.mlr.press/v28/wan13.html.

Zubek, Julian and Dariusz M. Plewczynski (2016). "Complexity curve: a graphical measure of data complexity and classifier performance". In: *PeerJ Computer Science* 2, e76. ISSN: 2376-5992. DOI: 10.7717/peerj-cs.76. URL: https://doi.org/10.7717/peerj-cs.76.